\documentclass[a4paper, conference]{IEEEtran}
\usepackage{cite}
\usepackage{graphicx}
\usepackage{amsmath}
\usepackage{amsmath}
\usepackage{amsfonts}
\usepackage{amssymb}
\usepackage{multirow}
\usepackage{xcolor}
\usepackage{float}
\usepackage{algorithm}
\usepackage{algorithmic}
\usepackage{caption}
\usepackage{subcaption}
\usepackage{url}
\begin{document}
\bstctlcite{IEEEexample:BSTcontrol}
%\RestyleAlgo{ruled} 

\title{Hierarchical Federated Learning for Crop Yield Prediction in Smart Agricultural Production Systems}

% \author{
% \hspace{-0.75cm}
% Hamza Reguieg$^1$, Mohammed El Hanjri$^2$, Mohamed El Kamili$^1$, Abdellatif Kobbane$^2$ \vspace{0.25cm}\\
% %, Jalel Ben-othman$^2$, Khalid Chougdali$^3$ \vspace{0.25cm}\\
% $^1$Higher School of Technology, Hassan II University in Casablanca, Morocco \\
% $^2$ENSIAS, Mohammed V University in Rabat, Morocco.\\
% %$^2$Université Sorbonne Paris Nord - CentraleSupélec, France,\\
% %$^3$ENSA, Ibn Tofail University, Kenitra, Morocco.
% \vspace{0.1cm}\\
% \textit {hamza.reguieg-etu@etu.univh2c.ma, mohammed.elhanjri@um5r.ac.ma, \\ abdellatif.kobbane@ensias.um5.ac.ma, mohamed.elkamili@univh2c.ma}}
% %}
\author{
\hspace{-0.75cm}
Anas Abouaomar$^1$, Mohammed El hanjri$^1$, Abdellatif Kobbane$^1$, Anis Laouiti$^2$, Khalid Nafil$^1$\\
\\

$^1$ENSIAS, Mohammed V University in Rabat, Morocco\\
$^2$ Samovar, Télécom SudParis, Institut Polytechnique de Paris, France\\
\\
\vspace{0.1cm}
\textit {anas\_abouaomar@um5.ac.ma, mohammed.elhanjri@um5r.ac.ma, abdellatif.kobbane@ensias.um5.ac.ma, 
}\\
\textit{anis.laouiti@telecom\-sudparis.eu, khalid.nafil@ensias.um5.ac.ma}
}

\maketitle

\begin{abstract}

%\vspace{7cm}

In this paper, we presents a novel hierarchical federated learning architecture specifically designed for smart agricultural production systems and crop yield prediction. Our approach introduces a seasonal subscription mechanism where farms join crop-specific clusters at the beginning of each agricultural season. The proposed three-layer architecture consists of individual smart farms at the client level, crop-specific aggregators at the middle layer, and a global model aggregator at the top level. Within each crop cluster, clients collaboratively train specialized models tailored to specific crop types, which are then aggregated to produce a higher-level global model that integrates knowledge across multiple crops. This hierarchical design enables both local specialization for individual crop types and global generalization across diverse agricultural contexts while preserving data privacy and reducing communication overhead. Experiments demonstrate the effectiveness of the proposed system, showing that local and crop-layer models closely follow actual yield patterns with consistent alignment, significantly outperforming standard machine learning models. The results validate the advantages of hierarchical federated learning in the agricultural context, particularly for scenarios involving heterogeneous farming environments and privacy-sensitive agricultural data.

\end{abstract}

\IEEEpeerreviewmaketitle

\section{Introduction}

The agricultural sector is undergoing a crucial transformation as it faces the combined pressures of global population growth, climate change and increasing demand for food security. One of the fundamental elements of sustainable and efficient agricultural production is the accurate forecasting of crop yields. Forecasting crop yields enables farmers, agronomists and policy-makers to make data-driven decisions on resource allocation, supply chain planning, market pricing and food distribution. Traditional crop yield forecasting models rely on centralized machine learning techniques, which require the collection and aggregation of large amounts of sensor, satellite and historical yield data from multiple farms into a central database. While effective in controlled environments, these centralized approaches raise significant concerns in real-life farming contexts.

Firstly, agricultural data is naturally heterogeneous and geographically distributed. Farms vary considerably in terms of soil quality, climatic conditions, crop type, cultivation techniques and resource use. Collecting this data in a centralized system often requires high communication overheads and reliable connectivity, which is not always possible in rural or low-infrastructure areas \cite{9086620}. Secondly, privacy and data ownership are major concerns in modern agriculture. Farm owners are often reluctant to share sensitive operational data with third parties for competitive, ethical or legal reasons. These limitations underline the need for decentralized, privacy-preserving learning mechanisms capable of leveraging data from multiple farms without the need for data centralization.\\

Federated Learning (FL) has emerged as a promising solution to address these challenges. FL allows multiple distributed clients (e.g., farms, devices, or regions) to collaboratively train a shared global model under the coordination of a central server, while keeping local data on device. Only model updates (e.g., gradients or weights) are exchanged, ensuring data privacy and minimizing communication cost. FL has been successfully applied in healthcare, finance, water consumption, and IoT, and its applications in precision agriculture are gaining momentum \cite{chauhan2024review, el2023federated}. Recent studies have shown the potential of FL in tasks such as crop classification, soil analysis, pest detection, and yield forecasting.

Despite these advances, existing FL-based approaches in agriculture are often limited by static client participation, uniform model aggregation, and lack of adaptation to seasonal and crop-specific variations. In real agricultural systems, farms cultivate different crops across seasons, and their participation in collaborative learning may vary over time. Moreover, applying a single global model across all farms fails to capture the variability introduced by crop type, climate region, or local farming practices. These limitations can significantly limit the accuracy and generalizability of yield prediction models.\\

To address these challenges, we propose a novel Hierarchical Federated Learning architecture specifically designed for smart agriculture and crop yield prediction. Our approach introduces a crop-aware, seasonal subscription mechanism where each farm dynamically joins a crop-specific cluster at the beginning of each agricultural season. Within each cluster, clients collaboratively train a specialized model tailored to the targeted crop type. These cluster-level models are then aggregated by a central server to produce a higher-level global model that integrates knowledge across multiple crops. This dual-layer aggregation strategy improves model personalization, enhances generalization, and reflects the dynamic and heterogeneous nature of agricultural data.

The key contributions of this paper are as follows:
\begin{itemize}
    \item We design a seasonal and crop-type-clustered Federated Learning paradigm for smart agriculture, enabling dynamic client participation aligned with crop production cycles.
    \item We develop a hierarchical model aggregation process that balances local specialization (per crop) with global generalization across crop types.
    \item We demonstrate the effectiveness of the proposed system through comprehensive experiments.
\end{itemize}

The rest of this paper is organized as follows, with section II discussing related works. The system model is described in Section III, while Section IV presents the simulation and numerical findings. Section V serves as the paper's conclusion.\\

\section{Related Works}

Federated Learning has recently emerged as a compelling paradigm in smart agriculture, offering a privacy-preserving and communication-efficient alternative to centralized data analytics. Its decentralized nature makes it suitable for agricultural scenarios, where data is often generated at distributed farm-level sensors and subject to strict privacy concerns.

A performance evaluation of FL architectures in the context of crop yield prediction demonstrated that centralized FL tends to outperform decentralized versions in terms of accuracy and latency. This finding highlights the need for infrastructure-aware FL deployment strategies in farming ecosystems \cite{mukherjee2024federated}.

A comprehensive survey of FL techniques in agriculture identified key architectural patterns and categorized models into centralized, hierarchical, and fully decentralized. This work also discussed optimization challenges arising from non-IID data, the impact of aggregation strategies such as FedAvg and FedProx, and the need for interpretability in agricultural forecasting tasks \cite{hiremani2025federated}.

Ensemble learning strategies have shown promise in improving the accuracy of crop yield forecasting. A notable example is a precise expert system that combines multiple imputation, ant colony optimization for feature selection, and an Extra Trees classifier, achieving robust results even in noisy datasets. The modular design of this system provides a transferable foundation for future integration within federated settings \cite{tripathi2024design}.

Deep learning architectures have also been explored extensively. A CNN-RNN framework was used to predict corn and soybean yields, effectively modeling both spatial and temporal dynamics \cite{khaki2020cnn}. Similarly, a multi-modal fusion architecture incorporating hyperspectral and LiDAR data, combined with attention mechanisms, demonstrated substantial improvements in predictive accuracy for plant breeding applications \cite{aviles2024integrating}.

Hybrid models combining weather forecasts with remote sensing data were shown to significantly improve wheat yield prediction, particularly under variable climate conditions. The system integrated time-series learning, highlighting the relevance of incorporating temporal features in FL pipelines \cite{peng2024deep}.\\

FL was directly applied in a yield forecasting model using diverse datasets including soil profiles, weather data, and crop management parameters. The results revealed that federated training consistently outperformed local-only models in predictive accuracy, demonstrating FL's real-world feasibility in farming \cite{manoj2022federated}.

Advanced feature engineering and component analysis on vegetation indices led to the design of a DL-based seasonal prediction model for maize crops. Although implemented in a centralized setup, its architectural patterns are highly compatible with FL designs \cite{pham2022enhancing}.\\

To enhance performance across multi-region agricultural datasets, ensemble FL models were proposed to aggregate predictions from heterogeneous clients. This method demonstrated scalability and robustness, particularly when applied to crops with regional variability in climate and soil \cite{khin2023harvest}.

In addition to ensemble-based strategies, another promising approach for enhancing federated learning under heterogeneous data distributions \cite{10322899} is coalition formation based on weight similarity. In \cite{el2024efficient}, a coalition-driven FL scheme was proposed where clients are dynamically grouped based on the Euclidean distance between their local model weights. Each coalition aggregates its models using barycentric averaging before contributing to a global aggregation, improving convergence speed and robustness in the presence of non-IID data. 

Reducing communication overhead is critical in FL for edge deployment. One approach to tackle this was model pruning, which significantly reduced the model size while preserving forecasting accuracy. This technique makes FL viable on embedded systems and low-power devices common in agriculture \cite{li2024model}.

Explainability is another essential component for FL adoption in practice. Integrating Explainable AI into FL frameworks allows for better transparency and decision traceability, especially crucial in agricultural decision-making processes where outcomes can have significant economic impact \cite{lopez2024interplay}. Complementary work reviewed the interplay between XAI and FL in sensitive domains, reinforcing the importance of human-centric model design \cite{rahmati2025federated}.

Additional strategies for improving trust in federated models include the use of fuzzy rule-based systems, which improve the interpretability of predictions across varied agricultural inputs \cite{daole2024trustworthy}. Further contributions introduced geometry-aware interpretability methods for FL at the edge level, supporting user confidence and debugging capabilities in constrained environments \cite{enyejo2024interpreting}.

\begin{figure*}
    \centering
    \includegraphics[scale=0.3]{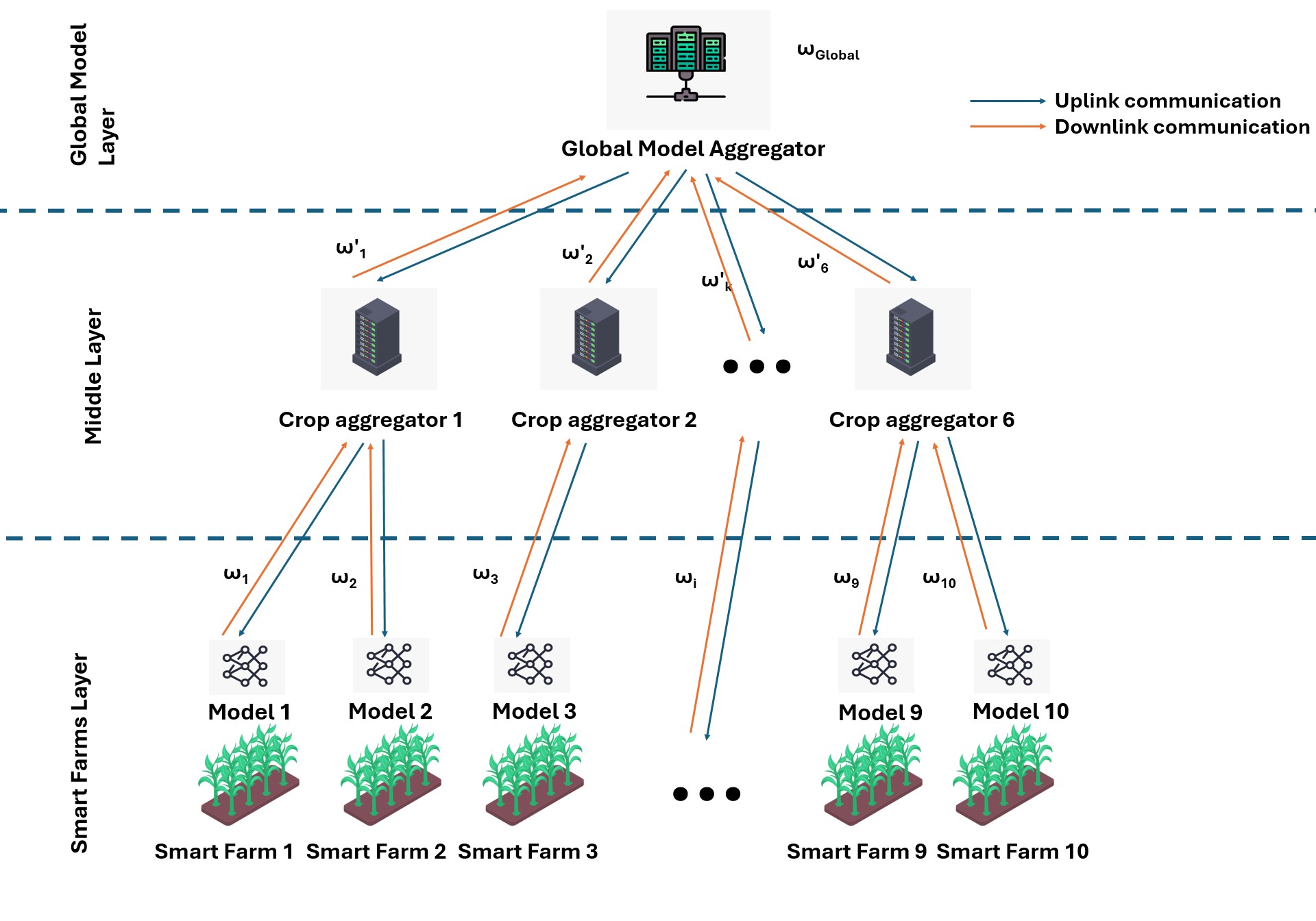}
    \caption{The architecture of the hierarchical federated learning model}
    \label{fig:1}
\end{figure*}

Innovations also emerged in graph-based learning, a hybrid federated model combining graph neural networks and recurrent architectures was developed to forecast crop yield, capturing both spatial relationships and temporal dependencies. This system demonstrated improved accuracy across highly diverse agricultural datasets \cite{nirosha2025enhancing}.

Finally, from an infrastructure perspective, the integration of FL with fog computing was explored as a means to reduce reliance on centralized cloud resources. This architecture is particularly relevant for agricultural applications in remote areas with limited connectivity \cite{vzalik2023review, abouaomar2019resources}.\\

These contributions collectively highlight the maturation of FL for agriculture. However, most implementations assume static client configurations and do not fully leverage the seasonal and crop-specific structure of agricultural production. The proposed work addresses this gap by introducing a dynamic subscription-based FL framework, where clients subscribe to crop-type clusters at the beginning of each season. This structure enables both specialized intra-cluster learning and the construction of a general global model, supporting both precision and scalability in yield prediction across heterogeneous farming contexts.\\

%\section{System model}

\section{System Model and Problem Formulation}

We consider a hierarchical federated learning system composed of a set of farms (clients), a set of crop-specific clusters, and a central server. The training process proceeds in a seasonal manner: at the beginning of each season, each farm subscribes to a crop-type cluster and contributes to the training of the crop-specific model, which is subsequently aggregated by the server to form a global cross-crop model.\\

Fig.1 shows the three layer hierarchical federated learning architecture for smart agriculture system. At the bottom level, individual smart farms start by subscribing to one of the crop types then train local ML models on their proprietary crop data, improving their local model without sharing their raw data. These updates are transmitted to the crop-specific aggregators in the middle layer, which perform crop specific aggregation. The Global Model Aggregator at the top layer receives these partially aggregated models and computes the final global model \( w_{\text{global}} \), which is then distributed back to all participating farms at the start of the next farming round. This hierarchical design reduces communication overhead, enables asynchronous participation, and allows farms to benefit from collective learning and crop specific insights while maintaining complete control over their agricultural data.

\subsection{Client Local Training}

Let \( \mathcal{C} \) denote the set of all participating farms, and \( \mathcal{D}_i \) the local dataset of client \( i \in \mathcal{C} \). Each dataset consists of feature-target pairs for crop yield prediction:  
\[
\mathcal{D}_i = \{(x_j, y_j)\}_{j=1}^{n_i}, \quad x_j \in \mathbb{R}^d, \ y_j \in \mathbb{R}.
\]

Each client performs local training over its dataset by minimizing a loss function \( \mathcal{L}_i \) on its local model \( w_i \) initialized from the cluster model \( \theta_k \). The local update is:
\[
w_i^{(t+1)} \leftarrow \text{LocalUpdate}(\theta_k^{(t)}, \mathcal{D}_i) = \theta_k^{(t)} - \eta \nabla \mathcal{L}_i(\theta_k^{(t)}),
\]
where \( \eta \) is the learning rate and \( t \) is the current communication round.

\subsection{Crop-Specific Model Aggregation (Middle layer)}

Each farm subscribes to a crop-specific cluster \( \mathcal{G}_k \subseteq \mathcal{C} \), for crop \( k \in \{1, \dots, K\} \). The crop-specific model \( \theta_k \) is obtained by aggregating the local models of all clients in \( \mathcal{G}_k \) after \( E \) local epochs using weighted averaging:
\[
\theta_k^{(t+1)} = \sum_{i \in \mathcal{G}_k} \frac{n_i}{N_k} w_i^{(t+1)}, \quad \text{where } N_k = \sum_{i \in \mathcal{G}_k} n_i.
\]

This process repeats over several rounds \( T_k \) for each crop-type cluster, leading to a specialized model for each crop.

\subsection{Cross-Crop Aggregation (Global Model)}

At the end of the season, the central server aggregates the cluster models \( \{\theta_k\}_{k=1}^K \) into a global model \( w_{\text{global}} \). This model captures cross-crop knowledge and is useful for farms transitioning to other crops in future seasons. The global aggregation is defined as:
\[
w_{\text{global}} = \sum_{k=1}^{K} \alpha_k \theta_k, \quad \text{with } \alpha_k = \frac{N_k}{\sum_{j=1}^K N_j}.
\]

\subsection{Objective Function}

The overall goal is to minimize a global objective function composed of local empirical risks across all clients:
\[
\min_{w} \sum_{k=1}^{K} \sum_{i \in \mathcal{G}_k} \frac{n_i}{N} \mathcal{L}_i(w), \quad N = \sum_{k=1}^{K} N_k.
\]

The training process alternates between local optimization at the client level and hierarchical aggregation at the cluster and global levels.

\subsection{Hierarchical Federated Learning for Smart Farms Algorithm}

This paper introduces a hierarchical federated learning algorithm that implements a three-tier aggregation strategy for agricultural yield prediction, consisting of client-level local training, crop-specific cluster aggregation, and global cross-crop model synthesis. The algorithm employs a seasonal subscription mechanism where farms dynamically join crop clusters $\{\mathcal{G}_k\}_{k=1}^K$, performing local updates $w_i^{(t+1)} \leftarrow \text{LocalUpdate}(\theta_k^{(t)}, D_i, E)$ using gradient descent with learning rate $\eta$. At the middle layer, crop-specific models are computed via weighted averaging $\theta_k^{(t+1)} = \sum_{i \in G_k} \frac{n_i}{N_k} w_i^{(t+1)}$, where $n_i$ represents local dataset size and $N_k$ is the total cluster size. The global aggregation combines cluster models using $w_{\text{global}} = \sum_{k=1}^K \alpha_k \theta_k$ with weights $\alpha_k = \frac{N_k}{\sum_{j=1}^K N_j}$ proportional to cluster participation. The objective function minimizes the global empirical risk $\min_w \sum_{k=1}^K \sum_{i \in G_k} \frac{n_i}{N} L_i(w)$, where $L_i$ represents the local loss function. This hierarchical design enables both intra-cluster specialization for crop-specific patterns and inter-cluster knowledge transfer, demonstrating superior performance compared to standard federated averaging approaches while maintaining differential privacy through localized gradient computations.\\

\begin{algorithm}[h]
\caption{Hierarchical Federated Learning for Smart Agriculture}
\label{algo}
\begin{algorithmic}[1]
\REQUIRE Clients \( \mathcal{C} \), crop clusters \( \{\mathcal{G}_k\}_{k=1}^K \), rounds \( T_k \), local epochs \( E \)
\STATE Initialize \( \theta_k^{(0)} \leftarrow \theta_{\text{init}} \) for all \( k \in \{1, \dots, K\} \)
\FOR{each crop cluster \( k = 1, \dots, K \)}
    \FOR{each round \( t = 0, \dots, T_k - 1 \)}
        \FOR{each client \( i \in \mathcal{G}_k \) \textbf{in parallel}}
            \STATE \( w_i^{(t+1)} \leftarrow \text{LocalUpdate}(\theta_k^{(t)}, \mathcal{D}_i, E) \)
        \ENDFOR
        \STATE \( \theta_k^{(t+1)} \leftarrow \sum_{i \in \mathcal{G}_k} \frac{n_i}{N_k} w_i^{(t+1)} \)
    \ENDFOR
\ENDFOR
\STATE Compute global model: \( w_{\text{global}} \leftarrow \sum_{k=1}^{K} \frac{N_k}{N} \theta_k \)
\RETURN Global model \( w_{\text{global}} \) and crop-specific models \( \theta_k \)
\end{algorithmic}
\end{algorithm}

\section{Simulation and Numerical Results}
\subsection{Simulation Setup and Data Preprocessing}
All simulations were performed on an \textit{ASUS TUF A15} equipped with an \textit{AMD Ryzen 7 6800H} processor running at 4.7~GHz, 16~GB of RAM, and an \textit{NVIDIA RTX 3070 Ti} graphics card.\\

The simulations were run using synthetic data extended from publicly available agricultural datasets for corn, wheat, cotton, rice, soybean, and barley.

\subsection{Algorithm Implementation and Parameters}

The experiment was implemented using PyTorch and TensorFlow. The primary objective was to compare the performance of the hierarchical FL architecture. 3 ML models were used at the bottom layer, Random Forest, XGBoost, and a LSTM-CNN model (Long Short Term Memory Convolutionnal Neural Network) model were employed for the local training.\\

Hyperparameter tuning is important for deep learning forecasting performance, but this paper only evaluates federated learning methods. This section examines how personalization impacts model efficiency. We first investigate whether the local training for individual clients or crop specific approach produces better results.\\

The results were obtained using the client distribution in fig.2, where clients subscribe randomly to a crop type among the $K=6$ crop types predefined, with the only condition being that there must be at least 1 smart farm per crop type considering the total number of smart farms is $N=10$.\\

Each round farms perform training over $E=10$ epochs, and $T_k=15$ rounds per crop specific model. This was done to assess the effectiveness of both the local model and the crop specific model.

\subsection{Discussion}

\begin{figure}
    \centering
    \includegraphics[scale=0.5]{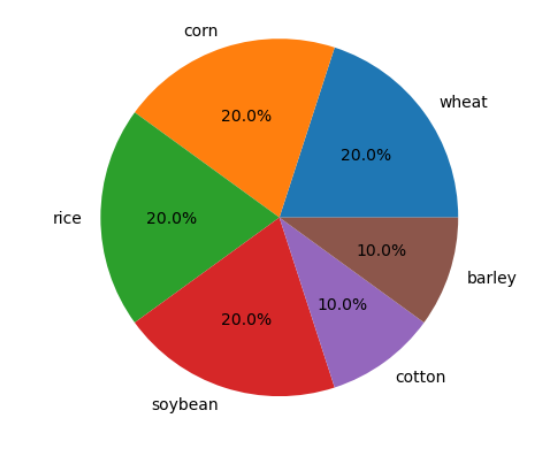}
    \caption{Number of clients per crop type}
    \label{fig:2}
\end{figure}

We chose 3 randomly selected smart farms each having subscribed to a different crop type (corn, wheat, and cotton). Fig. 3 shows the comparison between the different models predicted yield and the actual yield.\\

As shown in fig. (a), we can see that the local model and the crop specific model are able to make precise predictions about the farm yield. In contrast, the global model struggles to reach the same precision and in some cases makes inaccurate predictions which is similar to the standard ML model applied at large.\\

Fig.3 (b) and fig.3 (c) shows similar trends across all models with varying degrees of precision, the simulation results across all 3 smart farms show that both the local and crop specific models are able to predict the yield accurately, whereas the global model, much like standard ML models that don't take into account the nuances of specific crop dynamics, is often making very inaccurate predictions. This highlights the importance of our hierarchical federated learning model in helping farmers plan their smart farming operations.

\begin{figure}[htbp]
    \begin{subfigure}{0.3\textwidth}
        \centering
        \includegraphics[scale=0.4]{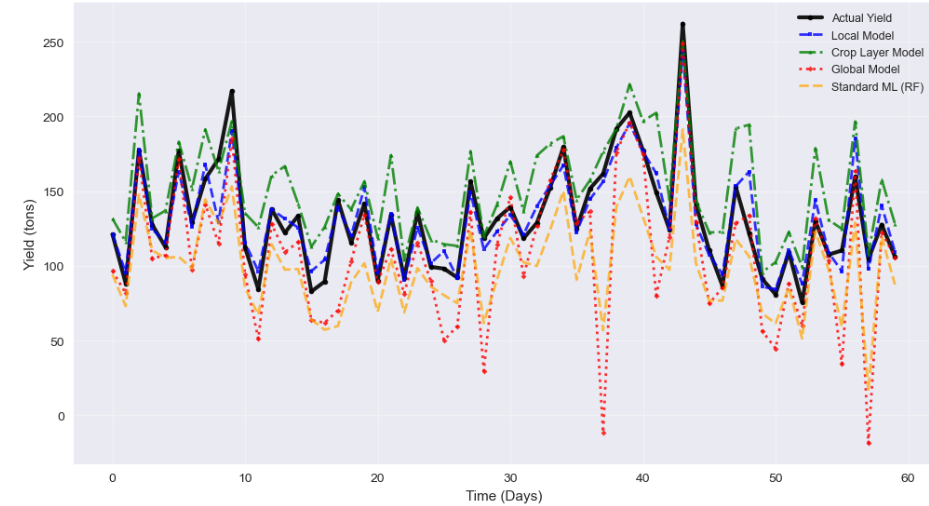}
        \caption{Comparison between the different models for Corn}
        \label{fig:plot1}
    \end{subfigure}
    \hfill
    \begin{subfigure}{0.3\textwidth}
        \centering
        \includegraphics[scale=0.4]{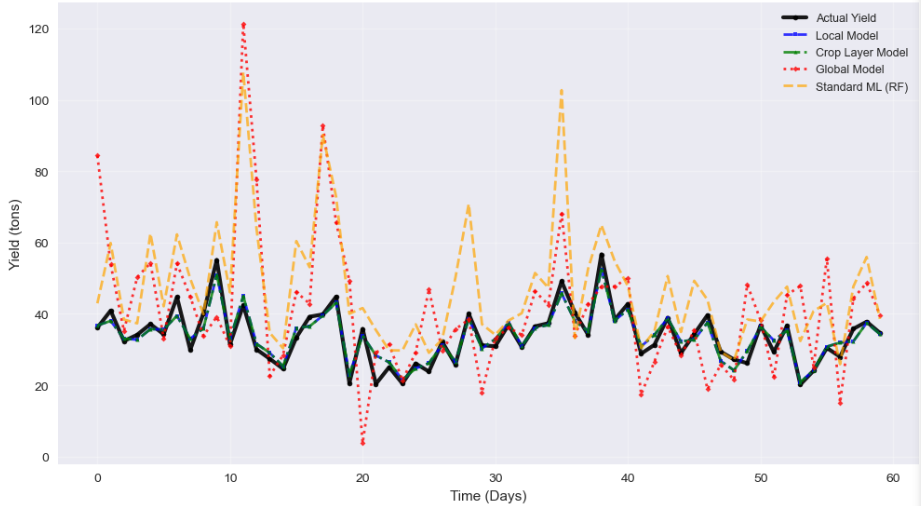}
        \caption{Comparison between the different models for Wheat}
        \label{fig:plot2}
    \end{subfigure}
    \hfill
    \begin{subfigure}{0.3\textwidth}
        \centering
        \includegraphics[scale=0.4]{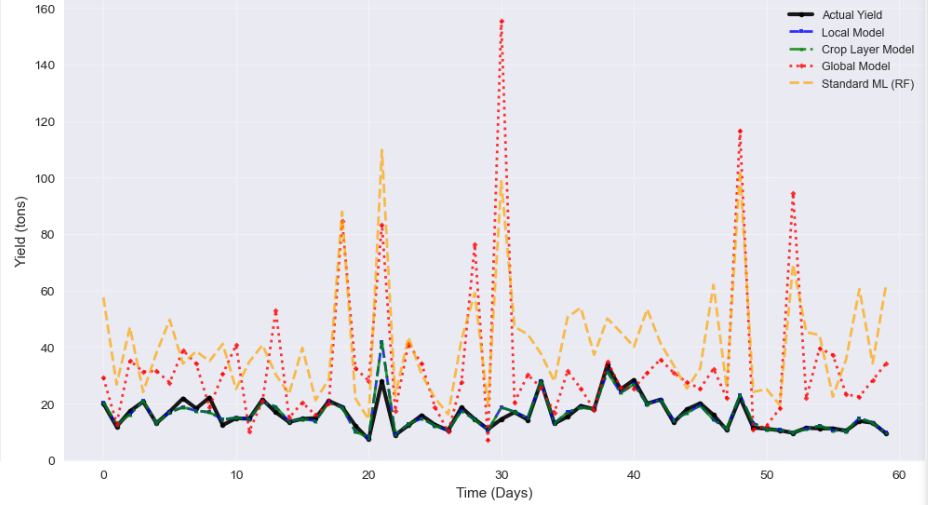}
        \caption{Comparison between the different models for cotton}
        \label{fig:plot3}
    \end{subfigure}
    \caption{Comparison between the different models for 3 chosen crops (Corn, Wheat, and Cotton)}
    \label{fig:three_plots}
\end{figure}

% \subsection{Simulation Setup}
% \subsection{Dataset}
% \subsection{Algorithm Implementation}
% \subsection{Model Architecture}
% \subsection{Parameters and Configuration}
%\subsection{Discussion}

\section{Conclusion}

In this paper, we presented a comprehensive hierarchical federated learning framework specifically tailored for crop yield prediction in smart agricultural production systems. The proposed three-layer hierarchical architecture successfully addresses key challenges in the smart agricultural context, including data privacy preservation, communication efficiency, and adaptation to the heterogeneous and seasonal nature of agricultural production. The dynamic seasonal subscription mechanism allows farms to join crop-specific clusters based on their current cultivation intentions, and the hierarchical aggregation strategy balances local specialization with global knowledge sharing. The experimental results demonstrate that both local and crop specific models achieve superior performance compared to both the global model and standard machine learning approaches. The hierarchical federated learning models maintain consistent results in comparison with actual yield trends while avoiding the volatility observed in the prediction made using traditional centralized models that are not able to distinguish the nuances between different crop types data.

Future works will focus on extending the system architecture by including more crop types, as well as exploring the viability of other clustering criteria such as, region based clustering, resource availability, or farm size.
\section*{Acknowledgement}

This research was supported by the PHC-Maghreb Project (grant number: 24MAG18/50188ZK).

\bibliographystyle{IEEEtran}
\nocite{*}
\bibliography{refs}

% that's all folks
\end{document}